\documentclass[letterpaper, 10 pt, conference]{ieeeconf}
\usepackage{times}
\IEEEoverridecommandlockouts % This command is only needed if you want to use the \thanks command
\usepackage[noadjust]{cite}

\usepackage{multicol}
\usepackage[bookmarks=true]{hyperref}

\usepackage{amsmath} 
\usepackage{amsfonts}
\usepackage{amssymb}
\usepackage{mathtools}
\usepackage{graphicx}
\usepackage{subcaption} 
\usepackage{float}
\usepackage{bm}

\usepackage[top=54pt, left=54pt, right=54pt, bottom=54pt]{geometry}
  
\usepackage[skip=3pt,font=footnotesize,labelfont=bf]{caption} 
\newcommand{\shrinka}{\def\baselinestretch{0.993}\large\normalsize}

\usepackage{cite}

\def\headervspace{\vspace{-3pt}}

\title{\vspace{-5pt}\LARGE \bf Learning Task Constraints from Demonstration \\for Hybrid Force/Position Control\vspace{-10pt}}
\author{Adam Conkey$^{1}$ and Tucker Hermans$^{1,2}$%
\thanks{$^{1}$School of Computing; Robotics Center; University of Utah, USA. $^{2}$NVIDIA, USA. \emph{Email: adam.conkey@utah.edu, thermans@cs.utah.edu}}}

\begin{document}
\shrinka
\maketitle
\shrinka

\begin{abstract}
  We present a novel method for learning hybrid force/position control from
  demonstration. We learn a dynamic constraint frame aligned to the direction of
  desired force using Cartesian Dynamic Movement Primitives. In contrast to
  approaches that utilize a fixed constraint frame, our approach easily
  accommodates tasks with rapidly changing task constraints over time. We
  activate only one degree of freedom for force control at any given time,
  ensuring motion is always possible orthogonal to the direction of desired
  force. Since we utilize demonstrated forces to learn the constraint frame, we
  are able to compensate for forces not detected by methods that learn only from
  demonstrated kinematic motion, such as frictional forces between the
  end-effector and contact surface. We additionally propose novel extensions to
  the Dynamic Movement Primitive framework that encourage robust transition from
  free-space motion to in-contact motion in spite of environment uncertainty. We
  incorporate force feedback and a dynamically shifting goal to reduce forces
  applied to the environment and retain stable contact while enabling force
  control. Our methods exhibit low impact forces on contact and low steady-state
  tracking error.
\end{abstract}

\IEEEpeerreviewmaketitle

\vspace{-2pt}
\section{Introduction}
\vspace{-2pt}
\label{sec:intro}

Many tasks, such as wiping a window, scrubbing a floor, and mixing in a bowl
require motion along a surface while maintaining a desired force. In order to
automate such constrained-motion tasks, robots must be able to control force and
position simultaneously. Though forces can be applied to an object using only
position control, it is generally unsafe to do so without force feedback since
excessively large forces can be imposed on the object (and the robot) in the
presence of estimation errors. Controlling forces relative to desired motion is
essential for performing constrained-motion tasks without risking damage to the
environment or the robot.

Hybrid force/position control is a popular control scheme for constrained-motion
tasks~\cite{raibert1981hybrid,ortenzi2017hybrid} since position and force
control objectives can be tracked simultaneously without conflict. Control is
performed with respect to a (possibly time-varying) Cartesian coordinate system
$\mathcal{C}_t \in SO(3)$, denoted the \textit{constraint frame}, that may be
arbitrarily located in space. Common choices for the constraint frame include
the world frame, the robot's tool frame, and frames attached to objects of
interest~\cite{siciliano2009robotics}. \textit{Task constraints} determine which
dimensions of the constraint frame are controlled for position and which are
controlled for force. They are typically defined by a diagonal binary selection
matrix $\mathbf{S}_t \in \mathbb{R}^{6\times 6}$ where $\mathbf{S}_t(i,i) = 1$
activates position control for the $i^{th}$ Cartesian dimension at time $t$ and
\(\mathbf{S}_t(i,i) =0\) enables force control.

\begin{figure}
  \centering
  \includegraphics[width=0.48\textwidth]{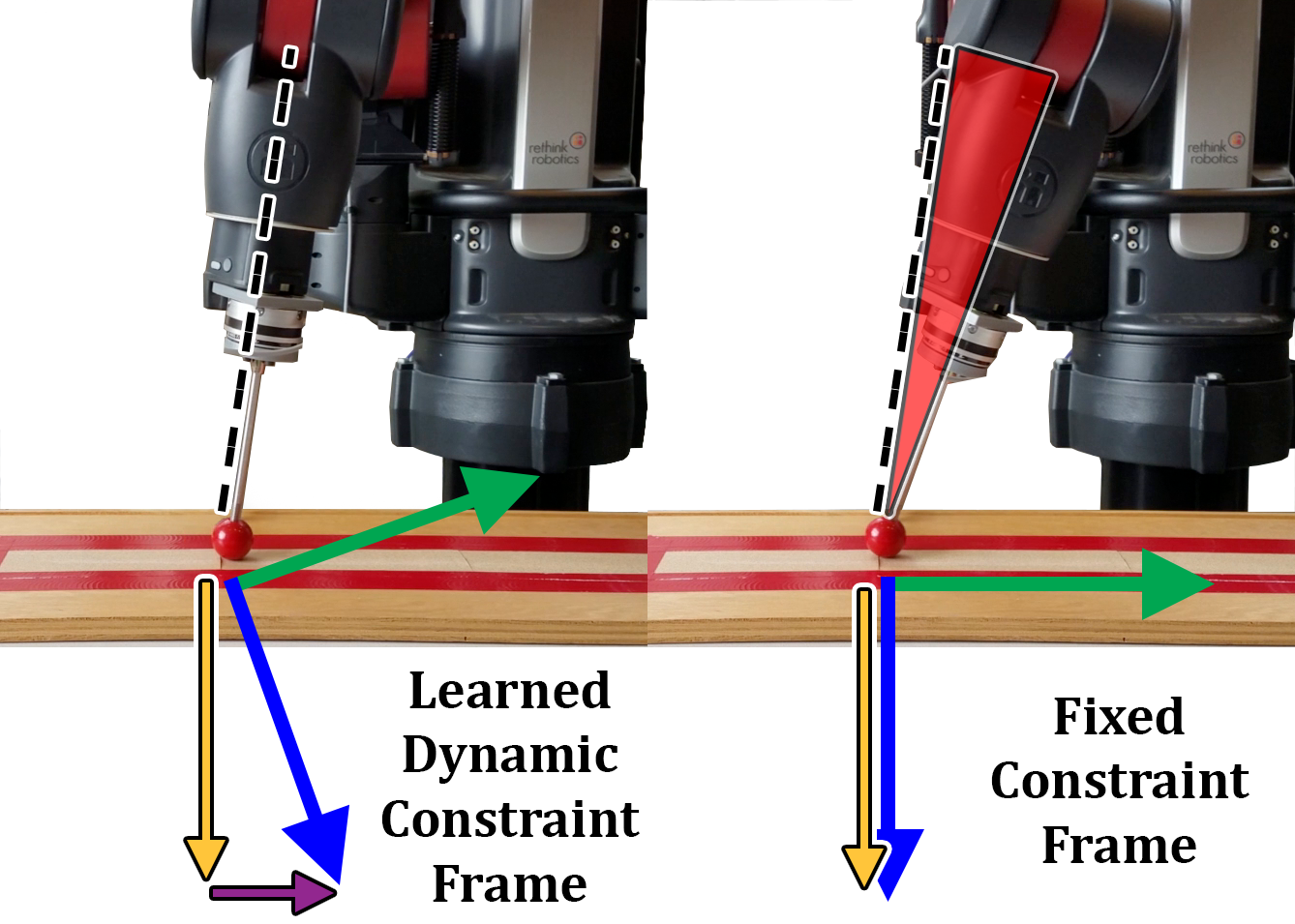}
  \caption{Illustration of a learned dynamic constraint frame (our method)
  versus a fixed constraint frame while sliding on a high-friction
  surface. Green and blue arrows indicate the $y$ and $z$ axes of the constraint
  frame, respectively, where force control is activated for the
  $z$-dimension. The fixed-frame approach tracks the desired force applied to
  the surface (yellow arrow), but incurs large pose error (red triangle) due to
  inhibited motion from frictional forces between the end-effector and contact
  surface. Our method actively compensates for these forces (purple arrow),
  thereby tracking both desired force (yellow arrow) and desired pose (dotted
  black line).}
  \label{fig:cover}
\end{figure}

Specifying an appropriate constraint frame and task constraints is difficult and
prone to error for complex tasks. Improper constraint specification is
especially problematic for transitioning from free-space motion to being in
contact with a surface, as large forces may be applied to the surface if
constraints are enabled too soon or too late~\cite{mandal1993experimental,
wilfinger1992comparison}. Even when constraints are properly specified, small
perturbations in the environment configuration or perceptual estimation errors
can interfere with the timing of the constraints. It is therefore desirable for
a robot to learn the constraints of a task and to adapt them to environment
uncertainty online.

While learning from demonstration has proven successful for learning task
constraints~\cite{peternel2017method, rozo2014learning,
steinmetz2015simultaneous, ureche2015task}, existing approaches focus on
learning axis-aligned constraints with respect to a chosen fixed frame. This is
a limitation when desired forces are time-varying and span multiple dimensions
of the constraint frame, as the robot loses a degree of freedom for motion with
each degree of freedom devoted to force control. Time-varying task constraints
have been learned from kinematic motion for the purpose of generalizing null
space policies~\cite{armesto2018constraint}. However, learning task constraints
only from kinematic motion leaves out valuable information contained in forces
observed during demonstration, such as frictional forces between the robot's
tool and the contact surface that need to be compensated for.

In this paper, we present a novel approach to learning task constraints for
hybrid force/position control from demonstration. We learn a dynamic constraint
frame using Cartesian Dynamic Movement Primitives (CDMPs) such that a principal
axis of the constraint frame is always aligned to the direction of desired
force. Our approach has the following benefits over existing methods that
utilize fixed constraint frames and that learn only from kinematic motion:
\begin{enumerate}
  \item \textbf{We accommodate tasks with constraints that change rapidly over
  time.} We show experimentally on the task of mixing in a bowl, in which forces
  vary in a non-trivial manner across all three dimensions of commonly used
  fixed frames, that we track both motion and force objectives. The task was not
  achievable controlling with respect to a fixed constraint frame.
  \item \textbf{We activate only one degree of freedom for force control at any
  given time.} Our method ensures motion is always possible orthogonal to the
  direction of desired force, whereas fixed-frame methods can render orthogonal
  motion impossible if all degrees of freedom need to be activated for force
  control.
  \item \textbf{We compensate for frictional forces between the end-effector and
  the contact surface while sliding.} Methods that learn task constraints only
  from kinematic motion neglect crucial information contained in observed
  forces, such as frictional forces between the end-effector and contact
  surface. We show experimentally on the task of sliding on a high-friction
  surface that frictional forces induce dramatic pose error when controlling
  with respect to a fixed constraint frame, while our method compensates for
  these forces and tracks desired force with little pose error (see
  Figure~\ref{fig:cover}).
\end{enumerate}

We additionally extend the Dynamic Movement Primitive (DMP) framework to
encourage robust transition from free-space motion to constrained motion. Our
extensions incorporate force feedback and contact awareness to reduce contact
forces and gradually transition into tracking desired forces. We also define a
dynamically changing goal that transitions as a function of the robot's contact
with the environment. These modifications account for cases in which the surface
to be contacted is not exactly at the anticipated position, e.g. due to
perceptual error or perturbation of the environment.

We structure the remainder of the paper as follows. We review related work in
the areas of learning force/position control and Dynamic Movement Primitives in
Section~\ref{sec:related_work}. In Section~\ref{sec:background}, we present the
base methods from the prior art we utilize in our framework. The details of our
novel contributions are provided in Section~\ref{sec:methods}. We describe our
experimental setup in Section~\ref{sec:experiment_setup} and present the
associated results in Section~\ref{sec:results}. Section~\ref{sec:discussion}
concludes with a brief discussion of our methods and directions for future work.

\headervspace
\section{Related Work}
\headervspace
\label{sec:related_work}

We review two general areas of related research: learning simultaneous control
of force and position, and incorporating force feedback into Dynamic Movement
Primitives.

\headervspace
\subsection{Learning Force/Position Control}
\headervspace

The literature in learning from demonstration for simultaneous control of
position and force has focused on 1) learning which dimensions of the constraint
frame should be selected for position or force control~\cite{deng2016learning,
peternel2017method,suomalainen2016learning, ureche2015task} and, to a lesser
extent, 2) learning the best constraint frame to control with respect
to~\cite{rozo2014learning, ureche2015task}. A key insight that has motivated
constraint selection methods is that dimensions of the constraint frame that
consistently exhibit high variance in force and low variance in position should
favor force control, and position control
otherwise~\cite{ureche2015task}. In~\cite{ureche2015task}, a criterion based on
trajectory variance is defined that modulates a stiffness parameter of a
Cartesian impedance controller, allowing force tracking when stiffness is
low. Impedance stiffness is set to zero in~\cite{suomalainen2016learning} for
compliant dimensions orthogonal to the dimension of highest variance in
motion. A series of boolean checks in~\cite{peternel2017method} over force and
position variance determines which axes of the robot's tool frame is enabled for
PI force control or Cartesian impedance control. In~\cite{deng2016learning},
binary constraint selection for a hybrid force/position controller is made by
enabling position control when the computed position variance is found to be
greater than the force variance.

Constraint frames are often chosen manually based on the requirements of the
task~\cite{raibert1981hybrid}. Common choices include the world
frame~\cite{hazara2016reinforcement, suomalainen2016learning}, surface
normals~\cite{deng2016learning}, the tool
frame~\cite{armesto2018constraint,peternel2017method, racca2016learning,
steinmetz2015simultaneous}, and frames attached to objects of interest in the
environment~\cite{ureche2015task}. The robot selects an appropriate constraint
frame from a collection of pre-defined candidate frames in~\cite{ureche2015task}
based on trajectory variance observed over multiple
demonstrations. In~\cite{rozo2014learning}, candidate frames include the start
and end frames of a human-robot collaboration task, and an appropriate frame is
chosen as the motion progresses. However, methods that use a fixed constraint
frame cannot be used for tasks in which desired forces span all three dimensions
of the constraint frame, as they require all dimensions to be enabled for force
control, thereby preventing simultaneous motion. A careful choice of constraint
frame can mitigate this problem, but for tasks in which desired forces vary in a
complex manner, fixed frame selection is infeasible.

Estimating task constraints and null space projections thereof can be used for
generalizing a task to different environment
configurations~\cite{armesto2018constraint,lin2017learning}. The task constraint
matrix and null space projection are estimated from motion data
in~\cite{lin2017learning} and incorporated into an operational space controller,
but task constraints for the purpose of force control are not
considered. In~\cite{armesto2018constraint}, the robot estimates task
constraints to command a policy learned from demonstration in the task null
space. While~\cite{armesto2018constraint} uses a force/torque sensor to align
the robot end-effector to the normal of a curved surface for generalizing a
learned planar task, it does not consider explicit task constraints for force
control and assumes the robot is already in contact with the surface before
initiating the task. Amanhoud et al.~\cite{amanhoud2019dynamical} use dynamical
systems to retain contact after a disturbance, but assume a known surface model
and control forces implicitly using impedance control.

\headervspace
\subsection{Force Feedback for Dynamic Movement Primitives}
\headervspace

Dynamic Movement Primitives (DMPs) are a widely used policy representation for
robot motion that afford real-time obstacle avoidance~\cite{park2008movement},
dynamic goal changing~\cite{pastor2009learning}, and can be learned from
demonstration using standard regression
techniques~\cite{ijspeert2013dynamical}. Various features of DMPs have been used
to augment motion trajectories with force information. Kormushev et
al.~\cite{kormushev2011imitation} synchronize position trajectories and force
profiles using the DMP phase variable. In~\cite{abu2015adaptation}, force error
is incorporated into the phase variable to aid in assembly tasks learned from
demonstration. Temporal coupling terms in~\cite{steinmetz2015simultaneous}
provide pose disturbance detection when executing tasks that repeatedly make and
break contact with a surface. Compliant Movement
Primitives~\cite{denivsa2016review} encode both motion and joint torques to
reduce contact forces during unexpected collisions. Velocity in periodic DMPs is
modulated based on a passivity criterion in~\cite{shahriari2017adapting} to
efficiently perform wiping tasks in a stable manner. Having both motion
trajectories and force profiles encoded as DMPs allows standard reinforcement
learning methods such as
%Policy Improvement with Path Integrals
$\text{PI}^2$ to be readily applied in order to learn the optimal forces needed
for completing a task~\cite{hazara2016reinforcement, kalakrishnan2011learning}.

Kober et al.~\cite{kober2015learning} learn DMPs for individual segments of a
multi-phase task and achieve force and position tracking with a hybrid
force/position controller. However, \cite{kober2015learning} selects a fixed
constraint frame based on convergence metrics of the DMPs, whereas our method
uses a dynamic constraint frame learned from forces observed during
demonstration. Several complementary works to ours use force information to
guide transitions between primitives
\cite{Kappler-RSS-15,kroemer2015towards,pastor2012towards}, but they do not
address the problem of robustly transitioning between free-space motion and
in-contact task phases. Steinmetz et al.~\cite{steinmetz2015simultaneous} handle
the case of ensuring contact when an expected contact is not satisfied, but
require switching between multiple controllers, which is known to suffer
stability
issues~\cite{driess2017constrained}. Additionally,~\cite{steinmetz2015simultaneous}
cannot adapt to contacts made sooner than expected. Our extensions to the DMP
framework enable robust transitions from free-space motion to constrained motion
using a single unified controller.

%%% Local Variables:
%%% mode: latex
%%% TeX-master: "root"
%%% End:

\headervspace
\section{Background}
\headervspace
\label{sec:background}

In this section, we present the base methods we employ in our framework. We
first define the hybrid force/position control law we use in
Section~\ref{sec:background:controller}, and then present a standard formulation
of DMPs in Section~\ref{sec:background:dmps}. Our novel contributions will be
presented in Section~\ref{sec:methods}.

\headervspace
\subsection{Hybrid Force/Position Controller}
\headervspace
\label{sec:background:controller}

We utilize the operational space hybrid force/position controller defined
in~\cite{khatib1987unified} which we present here for clarity. The controller we
use has the form
\begin{equation}
\label{eqn:hybrid_control}
\bm{\tau} = \bm{\tau}_f + \bm{\tau}_x + \left[ \mathbf{I}_n - \left(\mathbf{J}^\# \mathbf{J}\right)^T \right] \bm{\tau}_0 + \mathbf{g}
\end{equation}
where $\bm{\tau}_x$ and $\bm{\tau}_f$ are joint torques corresponding to
position and force control laws, respectively, $\bm{\tau}_0$ is an arbitrary
joint space control law commanded in the null space of hybrid force/position
control, and $\mathbf{g}$ is gravity compensation in joint space. For $n$ robot
joints, $\mathbf{I}_n \in \mathbb{R}^{n \times n}$ is the $n$-dimensional
identity matrix, $\mathbf{J} \in \mathbb{R}^{6 \times n}$ is the analytic
Jacobian, and
$\mathbf{J}^\# = \mathbf{M}^{-1}\mathbf{J}^T\bm{\Lambda} \in \mathbb{R}^{n
\times 6}$ is the generalized Jacobian pseudo-inverse derived
in~\cite{khatib1987unified}, where $\mathbf{M}$ is the joint space inertia
matrix and
$\bm{\Lambda} = \left[\mathbf{J}\mathbf{M}^{-1}\mathbf{J}^T\right]^{-1} \in
\mathbb{R}^{6 \times 6}$ is the inertia matrix reflected into task space. We use
the null space projection to command a low-gain PD controller $\bm{\tau}_0$ in
joint space that tracks a desired posture keeping the robot away from joint
limits when possible.

We use a Cartesian inverse dynamics controller defined as
\begin{equation}
\label{eqn:position_control}
\bm{\tau}_x = \mathbf{J}^T \bm{\Lambda} \bm{\Omega}\left(\mathbf{K}_p (\bm{x}_d - \bm{x}) + \mathbf{K}_d (\dot{\bm{x}}_d - \dot{\bm{x}}) + \ddot{\bm{x}}_d - \dot{\mathbf{J}}\dot{\bm{q}}\right)
\end{equation}
where $\bm{x}_d, \dot{\bm{x}}_d, \ddot{\bm{x}}_d$ are desired Cartesian poses,
velocities, and accelerations, $\bm{x}, \dot{\bm{x}}$ are actual poses and
velocities, $\dot{\bm{q}}$ are joint velocities, and
$\mathbf{K}_p, \mathbf{K}_d \in \mathbb{R}^{6 \times 6}$ are positive
semi-definite gain matrices. $\bm{\Omega} = \bm{\Omega}(\mathbf{S})$ is a block
tensor transformation that selects for position control in the constraint frame:
\begin{equation}
\label{eqn:omega}
\bm{\Omega}(\mathbf{S}) =
\begin{bmatrix}
\mathbf{R}^T\mathbf{SR} & \bm{0} \\
\bm{0} & \mathbf{R}^T\mathbf{SR} \\
\end{bmatrix} \in \mathbb{R}^{6\times 6}
\end{equation}
for $\mathbf{R} = \mathbf{^\mathcal{C}R}_0$ the rotation matrix from the base to
the constraint frame $\mathcal{C}$, and $\mathbf{S}$ the selection matrix
defined in Section~\ref{sec:intro}. Note: $\mathbf{S}$ and $\mathbf{R}$
generally vary with time but we drop the subscripts for consistency with the
controller terms.

We control forces using the following PI control law
\begin{equation}
\label{eqn:force_control}
\bm{\tau}_f = \mathbf{J}^T \bm{\tilde \Omega}\left(\mathbf{K}_f (\mathbf{F}_d - \mathbf{F}) + \mathbf{K}_I \sum_{t - \Delta_t}^t (\mathbf{F}_d - \mathbf{F}) \right)
\end{equation}
where $\mathbf{K}_f, \mathbf{K}_I \in \mathbb{R}^{6 \times 6}$ are positive
semi-definite gain matrices, $\mathbf{F}_d, \mathbf{F} \in \mathbb{R}^6$ are
desired and actual forces, and $\Delta t$ is the window of error
accumulation. $\bm{\tilde \Omega} = \bm{\Omega}(\mathbf{\tilde S})$ is the force
control selection matrix where $\mathbf{\tilde S} = \mathbf{I}_6 -
\mathbf{S}$. Force tracking occurs for each dimension of $\mathcal{C}$ that has
$\mathbf{S}(i, i) = 0$. We achieve pure free-space motion by setting
$\mathbf{S} = \mathbf{I}_6$.

In Section~\ref{sec:results:contact:sdt} we experimentally compare against PI
force control with Integral Error Scaling
(IES)~\cite{wilfinger1992comparison}. This technique attenuates the integral
error when it opposes the desired direction of force in order to mitigate the
chance of the end-effector breaking contact with the surface. For IES, the
integral error term in Equation~\ref{eqn:force_control}
$\Delta\mathbf{F} = \mathbf{F}_d - \mathbf{F}$ switches to
$\Delta\mathbf{F} = \beta(\mathbf{F}_d - \mathbf{F})$ for $\beta \in [0,1)$ when
$\mathbf{F}_d - \mathbf{F} < 0$.

\vspace{-2pt}
\subsection{Dynamic Movement Primitives}
\vspace{-2pt}
\label{sec:background:dmps}
We learn DMPs for position trajectories and force profiles following the
formulation of~\cite{pastor2011online} characterized by:
\begin{align}
\tau\dot{v} &= \alpha_v(g - y) - \beta_v v - \alpha_v(g-y_0)s + \alpha_v f(s) \label{eqn:tf_sys_1} \\
\tau\dot{y} &= v \label{eqn:tf_sys_2} \\
\tau\dot{s} &= -\alpha_s s \label{eqn:phase_variable}\\
f(s) &= \frac{\sum_{i} w_i\Psi_i(s)}{\sum_{i} \Psi_i(s)} s \label{eqn:forcing_function} \\
\Psi_i(s) &= \exp{\left(-h_i(s - c_i)^2 \right)} \label{eqn:basis_function}
\end{align}
Equations~\ref{eqn:tf_sys_1} and~\ref{eqn:tf_sys_2} define a first order
critically damped dynamical system for an appropriate choice of
$\alpha_v, \beta_v \in \mathbb{R}$ where $y$ is the state variable being
tracked, $y_0$ is the initial state, $g$ is the goal, and $f(s)$ a forcing
function. Equation~\ref{eqn:phase_variable} specifies the evolution of a phase
variable $s$ that decouples the system from explicit
time. Equation~\ref{eqn:forcing_function} defines the forcing function as a
normalized linear combination of basis functions. We use Gaussian basis
functions in Equation~\ref{eqn:basis_function} parameterized by centers $c_i$
and widths $h_i$, as is common in the literature~\cite{pastor2011online}. Each
degree of freedom receives its own DMP which are synchronized by the common
phase variable.

For orientation trajectories, we learn Cartesian Dynamic Movement Primitives
(CDMPs) similar to~\cite{pastor2011online} but with the full quaternion error as
suggested in~\cite{ude2014orientation}:
\begin{align}
\tau \bm{\dot{\omega}} &= \alpha_\omega \delta(\mathbf{q}, \mathbf{q}_d) - \beta_\omega\bm{\omega} - \alpha_\omega \delta(\mathbf{q}_0, \mathbf{q}_d)s + \alpha_\omega\mathbf{f}(s) \label{eqn:quat_tf_sys_1} \\
\tau\dot{\mathbf{q}} &= \frac{1}{2}\bm{\omega} * \mathbf{q} \label{eqn:quat_tf_sys_2}
\end{align}
Equations~\ref{eqn:quat_tf_sys_1} and~\ref{eqn:quat_tf_sys_2} are analogous to
Equations~\ref{eqn:tf_sys_1} and~\ref{eqn:tf_sys_2} where we define the
difference function for quaternions $\mathbf{q}_i = (v_i, \mathbf{u}_i)$ as
\(\delta(\mathbf{q}_1, \mathbf{q}_2) = 2\log(\mathbf{q}_2 *
{\mathbf{\overline{q}_1}})\). Here $\mathbf{\overline{q}}$ denotes quaternion
conjugation and the quaternion product is
\begin{equation*}
\mathbf{q}_1 * \mathbf{q}_2 = (v_1v_2 - \mathbf{u}_1^T\mathbf{u}_2) + (v_1\mathbf{u}_2 + v_2\mathbf{u}_1 + \mathbf{u}_1 \times \mathbf{u}_2)
\end{equation*}

\headervspace
\section{Methods}
\headervspace
\label{sec:methods}

We now present the details of our novel contributions. We describe our novel
approach to learning task constraints with a dynamic constraint frame in
Section~\ref{sec:methods:task_constraints}, and our novel extensions to the DMP
framework in Section~\ref{sec:methods:contact_dmps} that allow for robust
transition from free-space to in-contact motion.

\headervspace
\subsection{Learning Time-Varying Task Constraints}
\headervspace

\label{sec:methods:task_constraints}
Instead of learning the selection matrix $\mathbf{S}_t$ for a fixed constraint
frame (as in,
e.g.~\cite{peternel2017method,ureche2015task,armesto2018constraint,deng2016learning,suomalainen2016learning}),
we learn a dynamic constraint frame $\mathcal{C}_t$ for which $\mathbf{S}_t$ can
be specified in a canonical way. Our key insight is that, at each timestep, we
can align a principal axis of the constraint frame to the direction of desired
force $\mathbf{F}_{d,t}$, thereby requiring only one degree of freedom for force
control. We set the $z$-axis\footnote{The choice of $z$ is arbitrary.} to be
axis-aligned to forces observed during demonstration with selection matrix
values $\mathbf{S}_t(2,2) = 0$ and $\mathbf{S}_t(i,i) = 1$ otherwise. This
corresponds to force control along the $z$-axis of the constraint frame and
position control on all other axes.

We create the input to the learning procedure from the forces observed during
demonstration by defining the $z$-axis at each time step to be the observed
force vector normalized to unit length. We construct the other axes by selecting
the end effector $y$-axis as a candidate orthogonal axis and use cross products
to create a valid right-handed coordinate system. We learn a single CDMP
(described in Section~\ref{sec:background:dmps}) from the constructed input data
using ridge regression. The output is a smoothly varying trajectory for
$\mathcal{C}_t$ with a $z$-axis that tracks the direction of desired force. We
obtain a smoothly varying estimate of the magnitude of desired forces
$||\hat{\mathbf{F}}_{d}||$ to be applied along the $z$-axis of $\mathcal{C}_t$
by learning a DMP from $||\mathbf{F}_d||$. Our method inherits the
generalization benefits of DMPs well known in the
literature~\cite{ijspeert2013dynamical}. Thus, any modulations applied to the
robot's motion (e.g. temporal modulation) can also be applied to the learned
constraint frame and desired forces, ensuring motion and force objectives remain
in sync.
 
We show in Section~\ref{sec:results:bowl} that controlling with respect to our
learned constraint frame tracks desired forces using one degree of freedom for
force control, even when desired forces span multiple dimensions of fixed
reference frames such as the world or tool frames. We also show in
Section~\ref{sec:results:sdst:high_fric} that we achieve compensation of
frictional forces while sliding without explicitly modeling frictional
properties of the robot or the environment. This improves upon the typical
hybrid force/position control paradigm that makes the simplifying assumption of
frictionless contact~\cite{siciliano2009robotics}. Previous approaches for
learning hybrid force/position control from demonstration
(e.g.~\cite{deng2016learning, kober2015learning, steinmetz2015simultaneous}) do
not discover these forces and rely on low-friction environments to demonstrate
their methods.

\begin{figure*}[t]
  \centering
  \begin{subfigure}[b]{0.77\textwidth}
    \includegraphics[width=\textwidth]{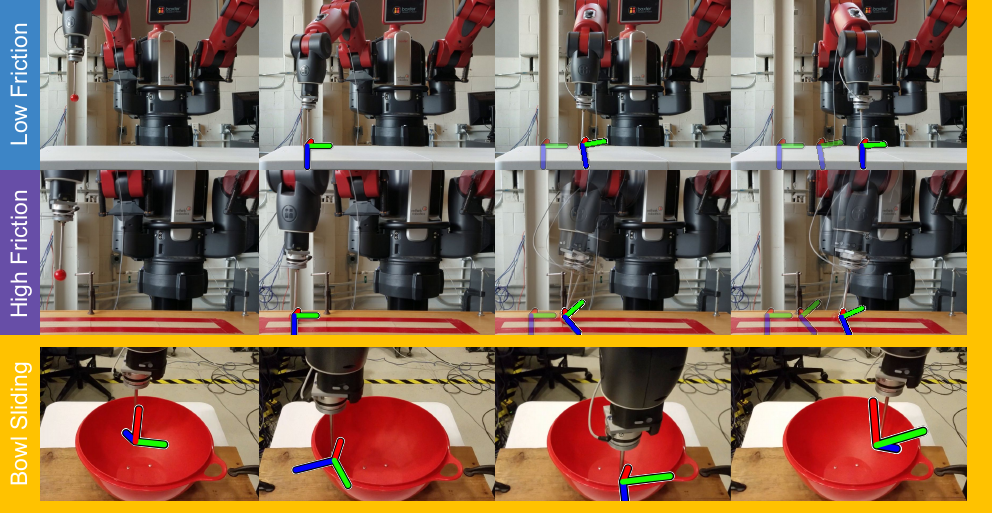}
    \vspace{-16pt}
    \caption{}
    \label{fig:sliding_seq}
  \end{subfigure}
  \begin{subfigure}[b]{0.17\textwidth}
    \includegraphics[width=\textwidth]{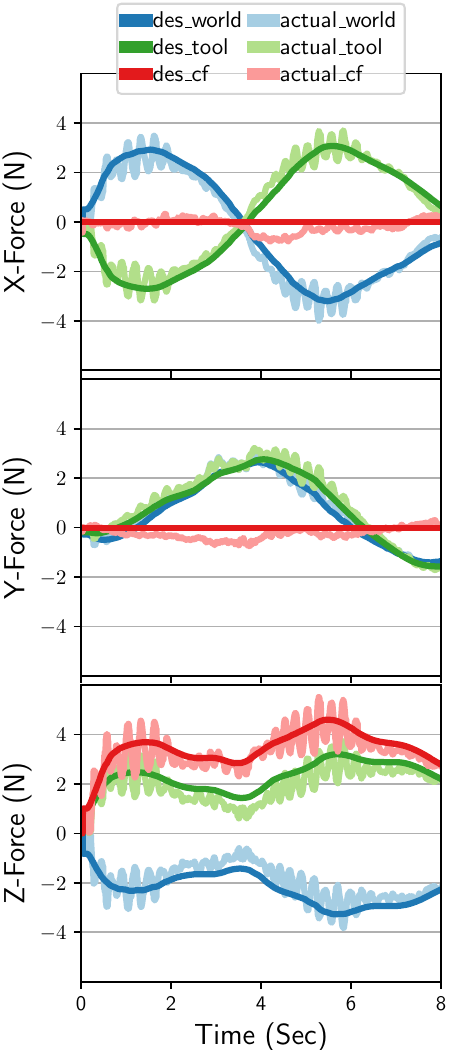}
    \vspace{-16pt}
    \caption{}
    \label{fig:bowl_forces}
  \end{subfigure}
  \caption{\textbf{(a):} Example scenarios performing hybrid force/position
  control with respect to a learned constraint frame (CF). We test sliding on
  low and high friction surfaces and a curved (bowl) surface. Coordinate frames
  show the pose of the learned CF over time (red = $x$, green = $y$, blue =
  $z$). Row~2 shows the difference in pose deviation for controlling with
  respect to the learned CF (primary image) and the world frame
  (semi-transparent overlay). \textbf{(b):} Comparison of desired and actual
  forces observed in the world, tool, and learned CFs while controlling with
  respect to the learned CF for the bowl scenario.}
  \label{fig:sliding}
\end{figure*}

\headervspace
\subsection{Extended DMPs for Making Stable Contact}
\headervspace
\label{sec:methods:contact_dmps}

We extend the DMP framework for the purpose of encouraging robust transition
from position control to force control when making contact with a surface.

\subsubsection{Halt DMP at Surface Contact}
\label{sec:methods:contact_dmps:halt}
To bring the system to a halt when the robot detects contact, we modify
Equation~\ref{eqn:tf_sys_2}:
\begin{equation}
\label{eqn:modified_tf_sys}
\tau\dot{y} = \frac{v}{1 + \alpha_f \sigma(f)|f| }
\end{equation}
where $f$ is the sensed force in the same task space dimension as $y$ and
$\alpha_f \in \mathbb{R}$ determines how sensitive the system is to contact
forces. We define the contact classifier $\sigma(f)$ as
\begin{equation}
\label{eqn:contact_classifier}
\sigma(f) = 
\begin{cases} 
1 & \mu_w > \mu_0 \\
0 & \text{otherwise} 
\end{cases}
\end{equation}
where $\mu_w$ is the mean value of $|f|$ over a sliding window of size $n$ and
$\mu_0$ is the mean of the noise inherent to the sensor. We show in
Section~\ref{sec:results:contact:sdt} that our method lowers impact forces when
contacting a surface earlier than anticipated.

The right-hand side of Equation~\ref{eqn:modified_tf_sys} has a similar form to
a term proposed in~\cite{ude2014orientation} for halting a DMP system when pose
error accumulates and in~\cite{abu2015adaptation} when force error
accumulates. However, in~\cite{abu2015adaptation} and~\cite{ude2014orientation}
the terms are applied to the phase variable and not the transformation
system. We apply our term directly to the transformation system velocity as it
allows us to selectively decouple the halting behavior of different
dimensions. We show in our experiments (Section~\ref{sec:results:contact:aat})
that the robot can halt motion in a dimension with an expected contact, while
the remaining unconstrained dimensions continue to converge to their desired
goal states. This cannot be achieved when the term is applied to the phase
variable, as it synchronizes control across all dimensions.

\subsubsection{Change in Goal Based on Contact Conditions}
\label{sec:methods:contact_dmps:change_goal}
If we assume the robot made the intended contact, but at an earlier time (see
Section~\ref{sec:discussion} for a further discussion of this assumption), then
the modification in Equation~\ref{eqn:modified_tf_sys} alone does not suffice
for completing the task. The DMP will remain in a halted state until the force
disappears, which will not happen when the force is due to an intended contact
and not a transient disturbance. We instead desire the free-space DMP system to
gracefully terminate its execution and transition into the in-contact phase of
the task. We achieve this by allowing the goal to dynamically change determined
by
\begin{equation}
\label{eqn:dynamic_goal}
\dot{g} = \alpha_c\sigma(f)(y - g) + \alpha_{nc} \left(1 - \sigma(f)\right) (g_0 - g)
\end{equation}
for $g_0$ the original goal, $g$ the current goal, $y$ the current DMP state,
and $\sigma(f)$ the contact classifier in Equation~\ref{eqn:contact_classifier}.

Equation~\ref{eqn:dynamic_goal} smoothly moves the current goal to coincide with
the robot's current state when the robot detects stable contact. Once the goal
and state coincide, the robot ends the free-space task phase and transitions to
the in-contact phase. If a disturbance caused the sensed force and it disappears
before the transition occurs, Equation~\ref{eqn:dynamic_goal} affords a smooth
transition back to the original goal and the phase proceeds from that point as
it would if no contact had been made.  Parameters
$\alpha_c, \alpha_{nc} \in \mathbb{R}$ control the rate of goal transition.

When the surface is farther than expected, the pose DMPs will converge to their
respective goals before making contact with the surface. At goal convergence,
each term in Equation~\ref{eqn:tf_sys_1} approaches zero, but we can still
incite movement toward the desired contact by moving the goal $g$ in the
direction of the desired contact by a small amount $\epsilon$. This moves the
end-effector at a constant velocity towards the desired contact, achieving
similar behavior to~\cite{steinmetz2015simultaneous,driess2017constrained}. Our
method improves over these methods as we do not require controller
switching~\cite{steinmetz2015simultaneous}, and we only require a single
demonstration as opposed to hundreds of real-robot
trials~\cite{driess2017constrained}.

\subsubsection{Incremental Force Control on Contact}
\label{sec:methods:contact_dmps:increment_fc}
When the robot first makes contact with the surface, an initial impact force
will be applied to the surface that depends on the velocity at impact; a higher
approach velocity results in a higher impact force. Though we mitigate impact
forces with the DMP feedback in Equation~\ref{eqn:modified_tf_sys}, we still
desire to enable force control when in contact in order to avoid sustained
application of high impact forces and to gracefully transition into the
constrained motion phase of the task. However, when the force error is large,
enabling force control instantaneously can make retaining stable contact with
the surface difficult, particularly for a stiff
environment~\cite{mandal1993experimental, wilfinger1992comparison}.

We propose to overcome this difficulty by incrementally enabling force control
for the desired dimension by leveraging the gradual goal transition of
Equation~\ref{eqn:dynamic_goal}. Instead of a strictly binary selection matrix
$\mathbf{S}_t$ for the hybrid force/position controller, we allow the Cartesian
dimension $i$ transitioning to force control to continuously vary from 1 to 0
determined by
\begin{equation}
\mathbf{S}_t(i,i) = \exp\left(1 - \frac{|y_c - g_c|}{|y - g|}\right)
\end{equation}
where $y_c$ is the system state at the time of contact, $y$ is the current
system state, $g_c$ is the DMP goal at the time of contact, and $g$ is the
current DMP goal. This expression equals 1 when the robot initially makes
contact, and converges to 0 as the goal $g$ converges to the current system
state $y$. This allows the controller to smoothly transition from position
control to force control as $\mathbf{S}_t(i,i)$ runs through convex combinations
of the two control laws. We show in Section~\ref{sec:results:contact:sdt} that
this affords stable contact and steady-state tracking when making contact at
different approach velocities.

%%% Local Variables:
%%% mode: latex
%%% TeX-master: "root"
%%% End:

\vspace{-2pt}
\section{Experimental Setup}
\vspace{-2pt}
\label{sec:experiment_setup}

We validate our methods on a Baxter robot equipped with a 6-axis Optoforce
HEX-E-200N force-torque sensor at the wrist. Both the robot state and the
force-torque sensor state\footnote{We use a 1.5Hz-cutoff online low-pass filter
for the force sensor.} are sampled at 1kHz. Robot controllers operate at
1kHz. The end-effector is a hard plastic sphere threaded to the tip of a steel
shaft which affords a point contact that can vary easily over the course of the
trajectory. Experiments were performed using an Intel Core i7-4770 CPU@3.40GHz-8
computer with 8GB of RAM running Ubuntu 14.04 and ROS Indigo. All software and
data is publicly available.\footnote{Data is available at
\texttt{\url{http://bit.ly/dmp_hfpc_data}}.}\footnote{Code is available at
\texttt{\url{http://bit.ly/dmp_hfpc_code}}.}\footnote{Video is available at
\texttt{\url{https://youtu.be/WzDP78K6ptI}}.}

We provide kinesthetic demonstrations by manually moving the robot arm in
gravity-compensation mode. Once recorded, the system autonomously segments the
demonstrations using the contact classifier in
Equation~\ref{eqn:contact_classifier} into three phase types: making-contact,
in-contact, and breaking-contact. Desired goal forces for making contact are
equal to the initial desired forces for the sliding phase. A DMP is learned for
each DOF in each task phase as described in Sections~\ref{sec:background:dmps}
and~\ref{sec:methods:task_constraints}. DMP parameters were set according to
guidance in the prior art~\cite{ijspeert2013dynamical}. All DMP and controller
parameters are kept the same in all experiments unless otherwise stated. We now
overview our experimental protocol; we present the associated results in
Section~\ref{sec:results}.

\begin{figure}[!b]
  \centering
  \includegraphics[width=0.44\textwidth]{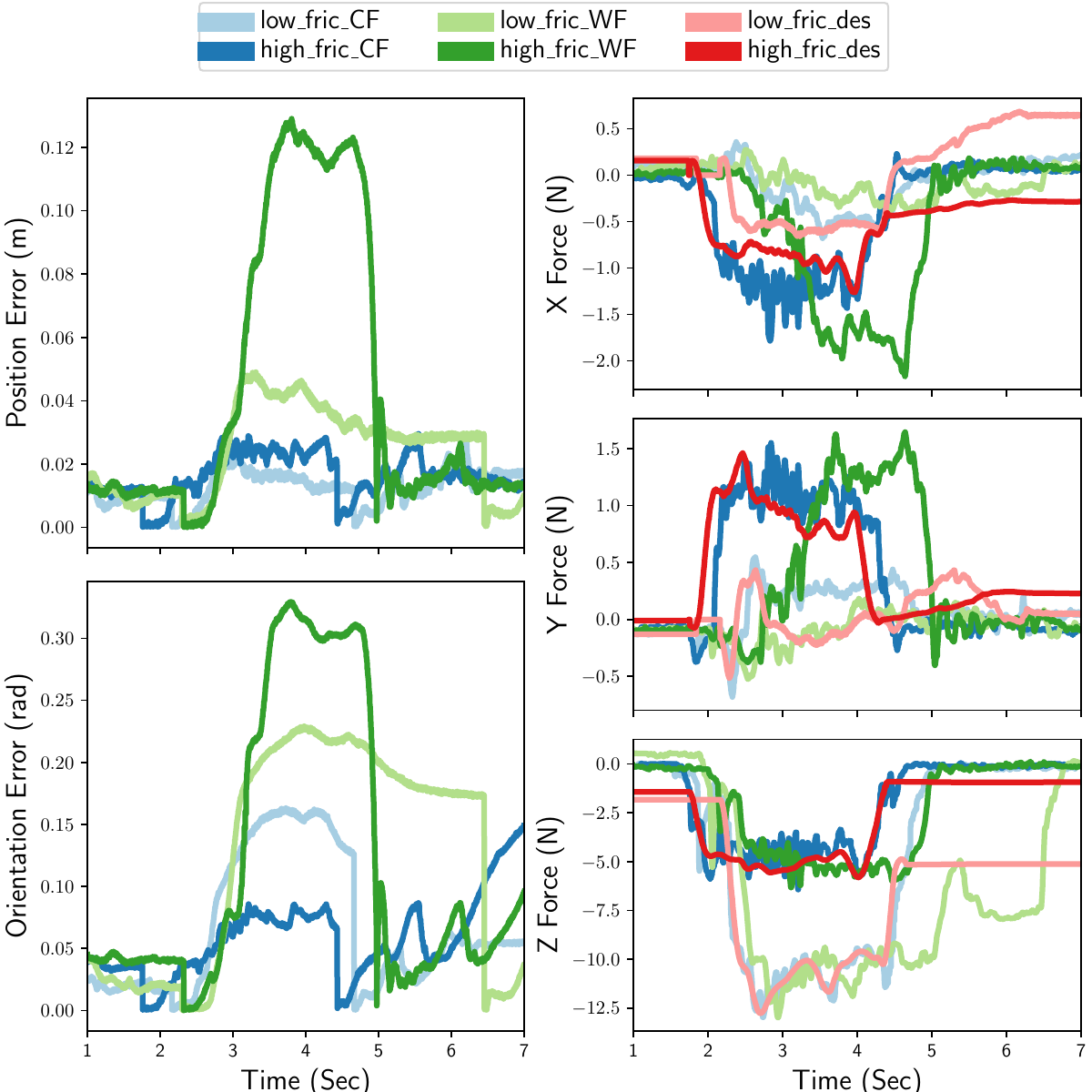}
  \caption{Results for sliding on low and high friction surfaces comparing control
  with respect to the world frame (WF) and our learned constraint frame (CF). The
  left plots show the L2 norm of the pose error of the tool over time. The right
  plots show the force profiles in the $x$, $y$, and $z$ dimensions.}
  \label{fig:sliding_compare}
\end{figure}

\headervspace
\subsection{Sliding on a flat surface.}
\headervspace We show that our learned dynamic constraint frame
(Section~\ref{sec:methods:task_constraints}) actively compensates for frictional
forces between the end-effector and contact surface while sliding. The
demonstrator slides the end-effector along the surface while applying a small
force and keeping the tool perpendicular to the table surface. We conduct
experiments across both low and high-friction surfaces and compare using our
learned dynamic constraint frame and the fixed world frame. Our results show
that controlling with respect to our learned constraint frame affords accurate
tracking of both the desired force profile and pose trajectory, while
controlling with respect to a fixed frame results in considerable pose error due
to frictional forces inhibiting the end-effector's motion.

\headervspace
\subsection{Sliding on a curved surface.}
\headervspace We demonstrate the ability of our learned constraint frame to
easily accommodate tasks where the constraints vary rapidly over time with the
task of performing a mixing motion on the interior curved surface of a
bowl. Desired forces for the task vary across all three dimensions of commonly
used frames such as world and tool frames, preventing simultaneous motion when
fixed-frame control approaches are utilized. Our method, on the other hand,
tracks the desired force profile and pose trajectory using only one degree of
freedom for force control.

\headervspace
\subsection{Making contact with a surface}
\headervspace We demonstrate the benefits of our methods presented in
Section~\ref{sec:methods:contact_dmps} for making robust contact with a
surface. We perform tests for making stable contact at varying table heights and
approach velocities, when given a single demonstration for moving the
end-effector straight down from an initial position above the table to a desired
contact point on the table at a nominal height. We compare against using the
standard DMP formulation with no force feedback (``open-loop'' below) and
against using PI force control with and without Integral Error Scaling. We also
consider the case where the end-effector moves at an angled approach to a
desired contact point on the table. These final experiments illustrate the
advantage of putting our DMP feedback on the transformation system instead of
the canonical system.

%%% Local Variables:
%%% mode: latex
%%% TeX-master: "root"
%%% End:

\begin{figure*}[t]
  \centering
  \begin{subfigure}[t]{0.7\textwidth}
    \includegraphics[width=\textwidth]{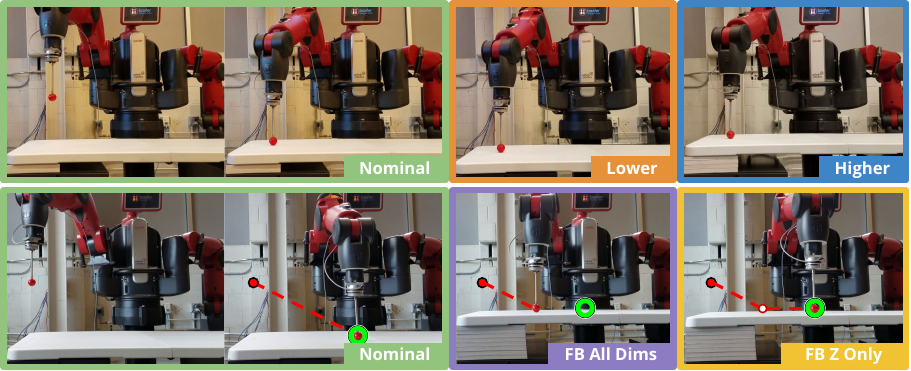} \vspace{-16pt}
    \caption{}
    \label{fig:contact_seq}
  \end{subfigure}
  \begin{subfigure}[t]{0.235\textwidth}
    \includegraphics[width=\textwidth]{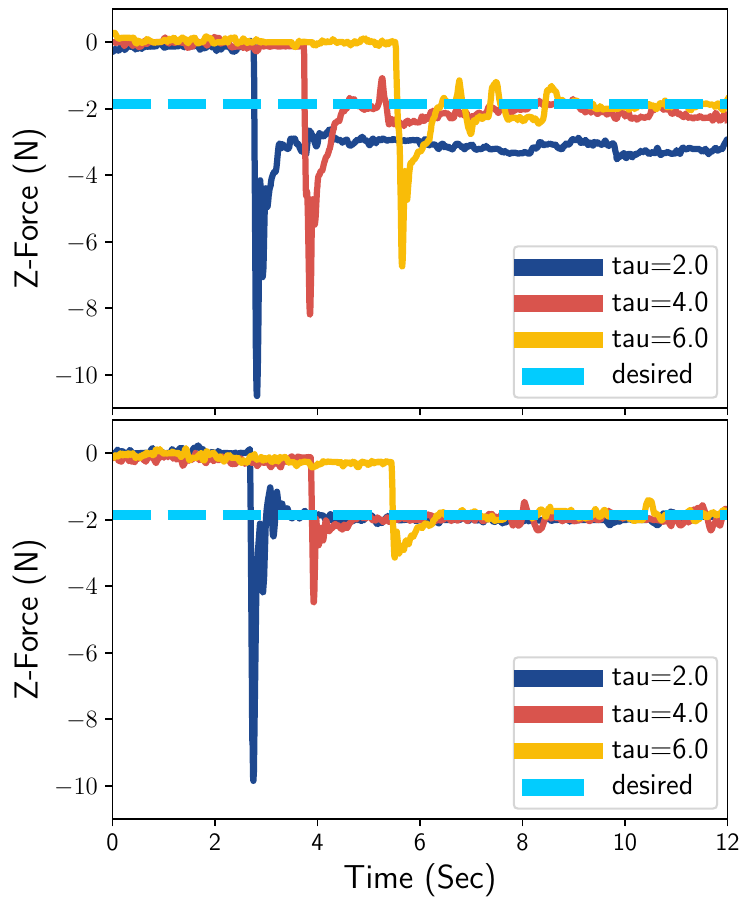}
    \vspace{-16pt}
    \caption{}
    \label{fig:hh_compare}
  \end{subfigure}
  \caption{\textbf{(a):} Scenarios we consider for robustly making contact. The
  top row illustrates making contact with a surface positioned at a different
  location than observed in demonstration. The bottom row illustrates making a
  slanted approach to a desired contact point and the effect of our DMP
  extensions when contact occurs sooner than anticipated. The dotted red line is
  the trajectory taken in each case. The green circle highlights the desired
  goal position. \textbf{(b):} The results for making contact sooner than
  expected at different approach velocities. The DMP temporal scaling parameter
  $\tau$ governs trajectory velocity, where lower values correspond to lower
  velocities. The dotted line shows the desired force. Top: IES force
  control. Bottom: our method.}
  \label{fig:contact}
\end{figure*}

\headervspace
\section{Results}
\headervspace
\label{sec:results}

We now present our experimental results following the same structure outlined in
Section \ref{sec:experiment_setup}.

\headervspace
\subsection{Sliding on a flat surface.}
\headervspace
\label{sec:results:sdst}
We perform the sliding task on two surfaces with drastically different friction
properties: a smooth plastic table and a piece of wood covered with sand
paper. In each case, the demonstrator slides the end-effector along the surface
while applying a small force and attempts to keep the end-effector oriented
perpendicular to the surface.

\subsubsection{Low-friction surface}
\label{sec:results:sdst:low_fric}
We learned a DMP trajectory and force profile from the provided demonstration
and executed it on the robot, as visualized in Figure~\ref{fig:sliding_seq}. We
compare our method of controlling with respect to a learned dynamic constraint
frame described in Section~\ref{sec:methods:task_constraints} against
controlling desired forces in the $z$-axis of the fixed world frame, which in
our setup is orthogonal to the table surface. Figure~\ref{fig:sliding_compare}
compares the resulting pose error and force profiles. The force profiles for
both methods are similar and adhere closely to desired forces. However, the
L2-norm of the pose error is noticeably higher for controlling with respect to
the world frame. This is due to unmodeled friction between the tool and table
inhibiting the sliding motion. Our learned constraint frame is less influenced
by this effect since it is aligned to the forces observed in demonstration,
including compensation forces for friction. See Figure~\ref{fig:cover} for a
diagrammatic visualization of the differences observed in controlling with
respect to learned and fixed constraint frames.

\subsubsection{High-friction surface}
\label{sec:results:sdst:high_fric}
We performed the same sliding experiment on a wooden board covered in 150 grit
sand paper. As seen in Figure~\ref{fig:sliding_seq}, the $z$-axis of the learned
constraint frame points primarily into the table where desired forces dominate,
but it also points slightly in the direction of motion. This is due to the
learned constraint frame aligning not only to the forces explicitly imposed by
the user during demonstration, but also the compensation forces the user was
implicitly applying to overcome friction and maintain the desired end-effector
orientation while sliding.

The pose error for controlling with respect to the world frame is exacerbated
further on the high-friction surface, as seen in
Figures~\ref{fig:cover},~\ref{fig:sliding_seq},~\ref{fig:sliding_compare}, and
our supplementary video. The pose error for controlling with respect to our
learned constraint frame remains low, and is in fact lower than for the smooth
table due to the overall lower forces being applied to the sand paper
surface. Figure~\ref{fig:sliding_compare} (right side) shows the force profiles
observed in each case. Both methods show good tracking in the $z$ dimension. Our
method also exhibits good tracking of the compensation forces for friction in
the $x$ and $y$ dimensions. We highlight that we achieve this without modeling
friction, and by using only one dimension of the constraint frame for force
control. Interestingly, the $x$ and $y$ forces for controlling with respect to
the world frame reach a similar magnitude, but at a delayed time. We suggest
this is due to frictional forces being passively reacted to, as opposed to being
actively commanded as our method does.

\headervspace
\subsection{Sliding on a curved surface.}
\headervspace
\label{sec:results:bowl}
A more complex force profile is achieved by sliding the end-effector along the
inside of a mixing bowl, as pictured in Figure~\ref{fig:sliding_seq}. We learn a
DMP trajectory and force profile and control the execution with respect to the
dynamic constraint frame learned from observed forces. We record the forces
observed during execution and transform them to the world and tool frames (both
commonly chosen constraint frames~\cite{siciliano2009robotics}) for the sake of
comparison. The recorded force profiles from one execution are visualized in
Figure~\ref{fig:bowl_forces}. We highlight that the desired forces
simultaneously vary in a non-trivial manner across all three dimensions of the
fixed world and tool frames. This implies that in order to track the forces with
these frames, all three dimensions would need to be activated for force control,
prohibiting simultaneous position tracking. Our method, on the other hand,
requires activation of only one degree of freedom for force control to track the
desired forces, thereby ensuring motion orthogonal to the direction of desired
force is always possible.

We attempted to compare against controlling with respect to a fixed constraint
frame. However, we were unable to perform the experiments safely. Enabling force
control for only one dimension, for example $x$ in the world frame, worked as
long as motion was primarily orthogonal to that direction. However, as soon as
motion started along that axis, control became unpredictable and had to be
terminated due to fears of damaging the robot and force sensor. Using a fixed
frame for this task requires very precise timing of constraint specification. We
will in future work seek a more reasoned criterion for determining this
specification.

\headervspace
\subsection{Making contact with a surface}
\headervspace
\label{sec:results:contact}

In our final experiments, we demonstrate the efficacy of our methods for making
robust contact presented in Section~\ref{sec:methods:contact_dmps} with the task
of making contact with a table for which the height may be higher or lower than
anticipated.
\subsubsection{Straight down approach}
\label{sec:results:contact:sdt}
We initialized the robot end-effector to hover 20cm in the world $z$-axis above
a table at a nominal height of 77cm measured from the ground to the table
surface. We recorded a demonstration that moved the end effector from its
initial position to a desired contact point on the table. The start and end
poses of the trajectory can be seen in Figure~\ref{fig:contact_seq}. We then
varied the height of the table to 74cm and 80cm. These heights were chosen to be
large enough to clearly illustrate the benefits of our methods while still
allowing for open-loop position trajectories to be executed for reference
without applying unsafe forces.

For the lower height of 74cm, open-loop position control leaves the end-effector
hovering approximately 3cm above the desired contact point. We use our DMP
extension described in Section~\ref{sec:methods:contact_dmps:change_goal} to
slowly change the goal in the direction of the desired contact. We chose a value
of $\epsilon = 0.0005$ to move the goal, as this value generates a slow enough
speed to easily make stable contact. Once the contact classifier detects
contact, the robot enables force control and tracks the desired initial force of
approximately 2N.

For making contact at the higher height of 80cm, we compare our method of DMP
force feedback with incremental force control against PI force control with and
without Integral Error Scaling (IES) described in
Section~\ref{sec:background:controller}. We test different execution speeds
using different values of the DMP temporal scaling parameter
$\tau \in [2, 4, 6]$, which approximately correspond to trajectory duration in
seconds. For each method we use the same control gains
$\mathbf{K}_f = \text{diag}(0.2), \mathbf{K}_I = \text{diag}(70.0)$ which were
empirically found to exhibit good steady-state tracking once already in contact.

We found that PI force control alone could not make stable contact at any speed
using these control gains; control immediately went unstable and had to be
terminated for safety. By introducing IES with a value of $\beta=0.001$, stable
contact was retained at each speed. However, as seen in the top of
Figure~\ref{fig:hh_compare}, there is steady-state tracking error of
approximately 1.5N for the case of $\tau=2.0$. The results for our method are
shown in the bottom of Figure~\ref{fig:hh_compare}. We achieve stable contact,
steady-state tracking, and reduce impact forces in all cases.

\subsubsection{Angled approach}
\label{sec:results:contact:aat}
Results for this case are pictured in Figure~\ref{fig:contact_seq}. The
end-effector was initialized to be approximately 25cm above a table height of
74cm. A demonstration was recorded that moved the end-effector at an angled
approach to the table along a straight line trajectory to a desired contact
point on the table. We compare two different behaviors possible with our DMP
feedback term defined in Equation~\ref{eqn:modified_tf_sys} on a table height of
86cm to make the difference apparent. When Equation~\ref{eqn:modified_tf_sys} is
activated for all task space dimensions (equivalent to applying the change on
the canonical system as previously proposed) using $\alpha_f = 10.0$, the
end-effector halts as soon as contact is detected and moves no further. The
end-effector reached the goal in the $z$-direction\footnote{Based on our change
in goal technique, the goal converges to the current pose when stable contact
has been retained sufficiently long.}, but cannot reach the $(x,y)$ goal even
though those directions are unconstrained. To achieve full goal convergence, we
activate Equation~\ref{eqn:modified_tf_sys} only for $z$, the dimension in which
contact is expected. In this case the end-effector makes contact, halts in the
$z$-direction but continues to converge to the position goal in other
directions.

%%% Local Variables:
%%% mode: latex
%%% TeX-master: "root"
%%% End:

\headervspace
\section{Discussion and Future Work}
\headervspace
\label{sec:discussion}

We presented a novel solution to learning hybrid force/position control from
demonstration. Our experimental results demonstrate that using a dynamic
constraint frame aligned to the direction of desired force allows
three-dimensional forces to be controlled accurately using only one degree of
freedom in the constraint frame. We additionally found that controlling with
respect to our learned constraint frame compensates for frictional forces
without any explicit modeling of friction, thereby reducing pose deviation over
controlling with respect to a fixed frame. An interesting avenue for future work
is to learn to adapt to surfaces with higher or lower friction than was observed
in demonstration. Reinforcement learning may be one promising approach to
achieve this sort of generalization \cite{hazara2016reinforcement}.

Our novel extensions to the DMP framework were shown to provide robust
transition from free-space to in-contact motion in spite of environment
uncertainty. Our method affords reduced impact forces and better steady-state
tracking on higher velocity impacts than other comparable methods. As indicated
in Section~\ref{sec:methods:contact_dmps:change_goal}, we assume an early
contact is the intended contact, as opposed to an undesired collision. We make
this assumption since the robot only uses a wrist force/torque sensor to
classify contacts. In most cases the robot could avoid observed obstacles using
collision avoidance techniques for DMPs~\cite{park2008movement}. When unintended
contact cannot be avoided, other perceptual modalities such as visual and
tactile feedback can allow for more robust classification of intended and
unintended contacts. We leave multi-sensory, robust contact classification as a
direction for future work.

%%% Local Variables:
%%% mode: latex
%%% TeX-master: "root"
%%% End:

\bibliographystyle{IEEEtran}
\vspace{-6pt}
\bibliography{references}

\end{document}